%% file: acl_latex.tex
\pdfoutput=1

\documentclass[11pt]{article}

\usepackage{acl}

\usepackage{times}
\usepackage{latexsym}

\usepackage[T1]{fontenc}

\usepackage[utf8]{inputenc}

\usepackage{microtype}

\usepackage{multirow}

\usepackage{hyperref}
\usepackage{algorithm}
\usepackage{algpseudocode}
\usepackage{amsmath}
\usepackage{multirow}
\usepackage[export]{adjustbox}
\usepackage{booktabs}
\usepackage{graphicx}
\usepackage{subcaption}

\title{Learning Open Domain Multi-hop Search Using Reinforcement Learning}

\author{Enrique Noriega-Atala \\
  The University of Arizona \\
  \texttt{enoriega@arizona.edu} \\\And
  Mihai Surdeanu \\
  The University of Arizona \\
  \texttt{ msurdeanu@arizona.edu} \\\And
  Clayton T. Morrison \\
  The University of Arizona \\
  \texttt{ claytonm@arizona.edu} \\}

\begin{document}
\maketitle
\begin{abstract}
  We propose a method to teach an automated agent to \emph{learn how to search} for multi-hop paths of relations between entities in an open domain. The method learns a policy for directing existing information retrieval and machine reading resources to focus on relevant regions of a corpus. 
  The approach formulates the learning problem as a Markov decision process with a state representation that encodes the dynamics of the search process and a reward structure that minimizes the number of documents that must be processed while still finding multi-hop paths. 
  We implement the method in an actor-critic reinforcement learning algorithm and evaluate it on a dataset of search problems derived from a subset of English Wikipedia. 
  The algorithm finds a family of policies that succeeds in extracting the desired information while processing fewer documents compared to several baseline heuristic algorithms.
\end{abstract}

\section{Introduction}
\input{content/introduction}

\section{Related Work}
\input{content/related_work}

\section{Learning to Search}

\input{content/focused_reading}


\section{Evaluation and Discussion}\label{sec:evaluation_wikification}

\input{content/experiments}
\section{Conclusions}
\input{content/conclusions}
\bibliography{anthology,custom}




\end{document}


\title{Suplementary Appendix}
\date{}
\maketitle

\section{Neural Network Architecture}
\begin{figure}[h]
\centering
  \includegraphics[width=0.5\textwidth]{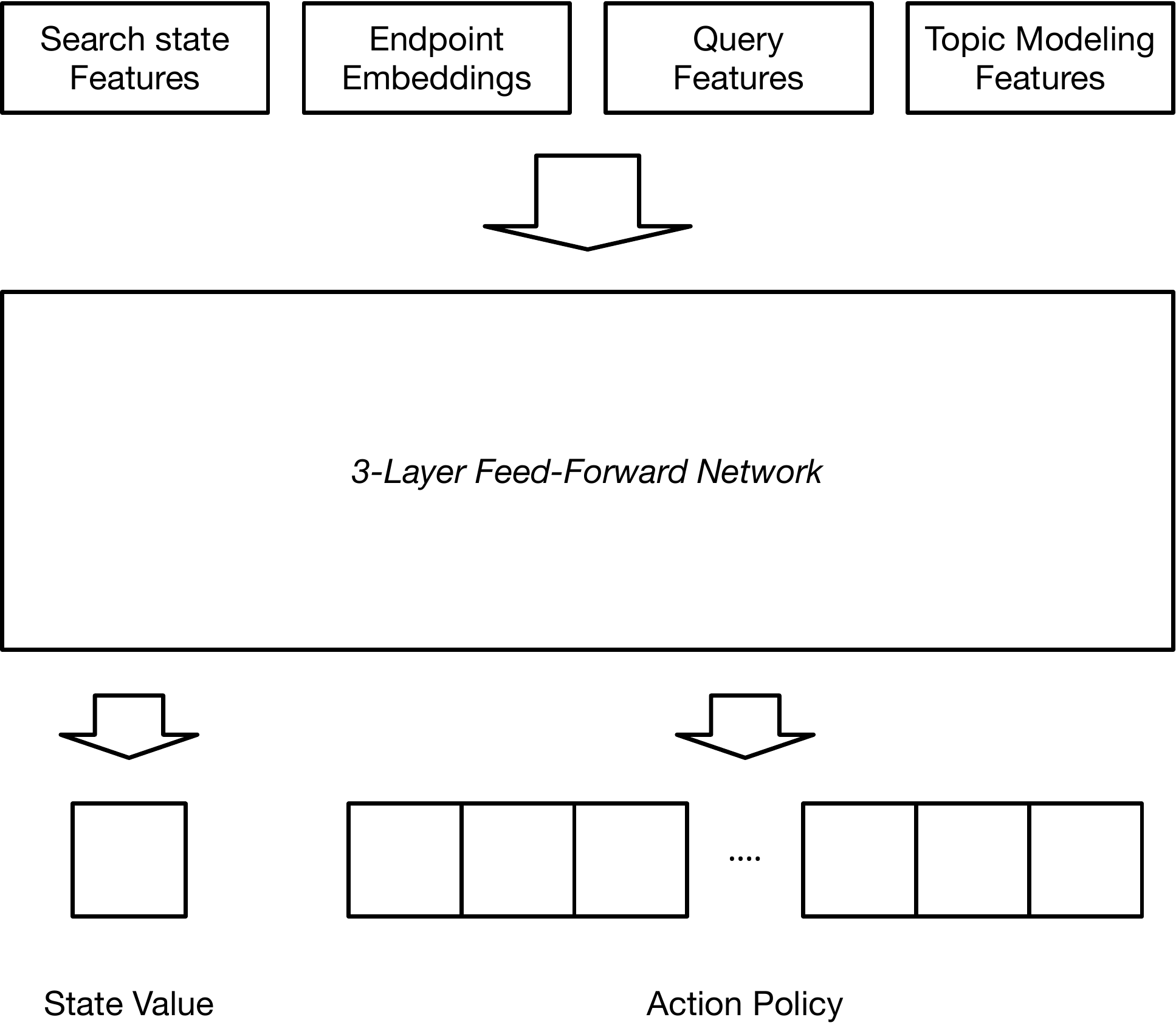}
  \caption{Neural Network Architecture}\label{fig:architecture}
\end{figure}

The neural network architecture used in our experiments is a standard fully connected, feed-forward network with four layers. Each layer has a tanh activation dropout, with coefficient $p=0.2$. Following the feed-forward layers, the architecture forks into two parallel output layers, the action policy, which has the same number of neurons as the action space, followed by a softmax activation and the state value layer, a single neuron with a linear activation. The endpoint embeddings are stacked with a dropout layer with coefficient $p$ left as a hyper-parameter. Figure \ref{fig:architecture} contains a schematic representation of the architecture. Table \ref{tab:architecture} lists the number of neurons and the total parameters of the architecture.

\begin{table}[hbt]
\centering
  \begin{tabular}{lr}
  \toprule
  \multicolumn{1}{c}{\textbf{\textit{Element}}} & \multicolumn{1}{c}{\textbf{\textit{Quantity}}}\\
  \midrule
    Layer 1 Neurons & 2100 \\
    Layer 2 Neurons & 1000 \\
    Layer 3 Neurons & 250 \\
    Layer 4 Neurons & 100 \\
    \midrule
    Action Policy Neurons & 15\\
    State Value Neurons & 1\\
    \midrule
    Total parameters & 3.79 mm\\
    \bottomrule
  \end{tabular}
  \caption{Neural network architecture quantities}\label{tab:architecture}
\end{table}

\section{Reinforcement Learning implementation}
We used the default parameters of the Advantage Actor Critic (A2C) implementation provided by \url{https://github.com/astooke/rlpyt}.
The library provides the RL algorithm implementation and a minibatch training algorithm. We used the default included values for A2C and for the optimizer (ADAM).

\section{Code and Data}
In order to preserve anonymity of this submission, we are not at this time disclosing the source code repository for this work. If the article is accepted for publication, we will post both elements in a website, along the detailed instructions on how to run the experiments to replicate the results.

%% file: content/introduction.tex
The sheer size of public corpora such as Wikipedia\footnote{\url{https://www.wikipedia.org/}} or large paper repositories like arXiv\footnote{\url{http://arxiv.org/}} and PubMed Central\footnote{\url{https://www.ncbi.nlm.nih.gov/pmc/}} poses an enormous challenge to automating effective search for relevant information.
This problem is compounded when the underlying information needs require {\em multi-hop connections}, e.g., searching for biological mechanisms that connect two proteins~\cite{cohen2015darpa} or searching for explanations that require complex reasoning by understanding text supported by different documents in QA systems~\cite{welbl-etal-2018-constructing,yang-etal-2018-hotpotqa}.

In a naive approach, an automated information extraction agent could process all the documents in a corpus, searching for the indirect connections that satisfy a multi-hop information need. However, this quickly becomes prohibitively expensive as the corpus size increases. Further, 
the documents may also be behind a paywall, adding an additional economic cost to accessing information. Thus, the naive exhaustive reading approach is simply not feasible for most large corpora scenarios. 
Instead, we need to incorporate the kind of iterative {\em focused reading} that humans are capable of. When people search for information, they use background knowledge, based in part on what they have just read, to narrow down the search space while selectively committing time and other resources to carefully reading documents that appear relevant. This process may be repeated multiple times until the information need is satisfied.

We propose a methodology that uses reinforcement learning (RL) to teach an automated agent how to direct a search process, using existing information retrieval and machine learning components selectively, focusing on the relevant parts of the corpus in order to minimize the expenditure of computational resources and access costs.

The contributions of our work are the following:
\begin{enumerate}
	\item A reinforcement learning framework to teach an automated agent how to direct a multi-hop search process that selectively allocates machine reading resources in an open-domain corpus.
	\item A set of domain-agnostic state representation features that enable the reinforcement learning method to learn a policy that improves the chances of finding the desired information while processing fewer documents compared to strong baselines.
	\item A new dataset of open-domain multi-hop search problems derived from English Wikipedia contained in the WikiHop dataset~\footnote{\url{http://qangaroo.cs.ucl.ac.uk}}~\cite{welbl-etal-2018-constructing}. 
	Using this dataset, we show that our RL approach is able to derive policies that find the desired information more frequently and by processing fewer documents than several heuristic baselines.
\end{enumerate}

%% file: content/related_work.tex
Modern machine reading technology enables the extraction of structured information from natural language data. Named-entity recognition~\cite{tjong-kim-sang-de-meulder-2003-introduction} systems detect and label specific classes of concepts from text, both in the general domain~\cite{manning} and for specific domains~\cite{neumann-etal-2019-scispacy}. Relation extraction systems extract interactions between different concepts in open-domain~\cite{DBLP:conf/tac/2018,DBLP:conf/tac/2017,DBLP:conf/tac/2008} and domain-specific scenarios~\cite{emnlp-2019-bionlp,ws-2019-bionlp,ws-2011-bionlp}.

Reinforcement learning has been successfully deployed for a variety of natural language processing (NLP) tasks. \cite{clark} proposed a policy-gradient method to resolve the correct coreference chains for the task of coreference resolution. \cite{li-etal-2017-end} used reinforcement learning to train an end-to-end task-completion dialogue system. For the task of machine translation, \cite{he2016dual} formulated the task as a dual-learning game in which two agents teach each other without the need of human labelers using policy-gradient algorithms.

Reinforcement learning has also been specifically applied to improving search and machine reading. In learning how to search, \cite{kanani2012selecting} proposed a methodology for the task of slot-filling based on temporal-difference q-learning that uses domain specific state representation features to select actions in a resource-constrained scenario. \cite{noriega-atala-etal-2017-learning} successfully applied RL to finding relevant biochemical interactions in a large corpus by focusing the allocation of machine reading resources towards the most promising documents. Similarly, \cite{wang2019automatic} explore the use of deep neural networks and deep RL to simulate the search behavior of a researcher, also in the biomedical domain.


%% file: content/focused_reading.tex
We propose a methodology to teach an automated agent how to selectively retrieve and read documents in an iterative fashion in order to {\em efficiently} find multi-hop connections between a pair of concepts (or entities). 
Each search step focuses on a restricted set of documents that are hypothesized to be more relevant for finding a connection between the two target concepts.
The focus set is retrieved and processed and if a path connecting the concepts is found, the search terminates. 
Otherwise, a new set of focus documents is identified based on what has been learned so far during the search. 
The process is repeated iteratively until the desired information is found or a number of iterations is exceeded.
Our method is general as it does not directly rely on any supervised domain specific semantics.

During the search, the agent iteratively constructs a {\em knowledge graph} (KG) that represents the relations between concepts found so far through machine reading.
In each iteration, the algorithm formulates a document retrieval query based on the current state of the knowledge graph, which is then executed by an information retrieval (IR) component. The IR component contains data structures to query the corpus, for example using an inverted index. The construction of these data structures usually only requires shallow processing, such as tokenization and stemming, and not a full-fledged NLP pipeline. 
Any documents returned from executing the query are processed by an information extraction (IE) component that performs named entity recognition and relation extraction.
The KG is expanded by adding newly identified entities as new nodes and previously unseen relations as new edges.
The overall goal of the method is to focus on the documents that appear to be most likely to contain a path between the target concepts, all while processing as few documents as possible.

	
\cite{noriega-atala-etal-2017-learning} formalized this iterative search process as a family of \emph{focused reading} algorithms, shown in Algorithm \ref{alg:focusedreading}:

%
%
%
%
\begin{algorithm}
\caption{Focused reading algorithm}
\begin{algorithmic}[1]
{\small
\Procedure{FocusedReading}{$E1,E2$}
   \State {\scriptsize $KG \gets \{\{E1, E2\}, \emptyset\}$} \label{alg:fr:graph}
   \Repeat \label{alg:fr:start}
   		\State {\scriptsize $Q \gets $ \Call{BuildQuery}{$KG$}} \label{alg:chquery}
   		\State {\scriptsize $(V, E) \gets$ \Call{Retrieval+Extraction}{$Q$}} \label{alg:ie}
   		\State {\scriptsize \Call{Expand}{$V, E,K G$}} \label{alg:reconcile}
   \Until{{\scriptsize \Call{IsConnected}{$E1,E2$}} OR {\scriptsize \Call{HasTimedOut}{}}}\label{alg:fr:stop}
\EndProcedure
}
\end{algorithmic}
\label{alg:focusedreading}
\end{algorithm}

The algorithm starts with the $KG$ representing only the \emph{endpoints} of the search: the named entities $E1$ and $E2$.
The algorithm then initiates the search loop. 
The first step in the loop analyzes the knowledge graph and generates an information retrieval query, $Q$. 
As we will describe shortly, the current state of the $KG$ is used to parameterize and constrain the scope of $Q$, focusing it on returning a limited subset of documents that are hypothesized to be most relevant. After retrieval, the documents are processed by the IE component. Any entities not previously found in the $KG$ are placed in the new entity set $V$, and similarly any new relations linking entities are placed in the new relation set, $E$. $V$ and $E$ are incorporated into the $KG$
and the algorithm then searches the updated $KG$ for any new possible paths connecting $E1$ and $E2$. If a path exists, it is returned as a candidate explanation of how $E1$ and $E2$ are related. 
Otherwise, if no such path exists, the query formulation process (using the updated $KG$) followed by IR and IE, is repeated until a path is found or the process times out.

This framework can answer multi-hop search queries for which the relationships along a connecting path come from different documents. For example, this process may discover that \emph{Valley of Mexico} is connected to the \emph{Aztecs} because the Aztecs were a \emph{pre-columbian civilization} (found in one document), which, in turn, was located in the Valley of Mexico (found in another document). 

In the following subsections, we formulate focused reading as a Markov decision process (MDP). 

\subsection{Constructing Query Actions}\label{sec:action_space}

%
%
%

\begin{table}[hbt]
\centering
\small
  \begin{tabular}{lll}
    \toprule
    \multicolumn{1}{c}{\textbf{\textit{Template}}}&\multicolumn{1}{c}{\textbf{\textit{\# Params}}}& \multicolumn{1}{c}{\textbf{\textit{Constraints}}}\\
    \midrule
    \emph{Conjunction}& Two: (A, B) & Contains $A$ and $B$\\
    \emph{Singleton}& One: (E) & Contains $E$\\
    \emph{Disjunction}& Two: (A, B) & Contains $A$ or $B$\\
    \bottomrule
  \end{tabular}
  \caption{Query templates}\label{tab:queries}
\end{table}

In the focused reading MDP, actions are comprised of information retrieval queries.
Actions are constructed from a set of three \emph{query templates}, listed in Table \ref{tab:queries}.
Each template is parameterized by one or two arguments representing the entities that are the subject of the query. The template type then incorporates these entities into the set of constraints that must be satisfied by a document in order to be retrieved.
The different query templates are intuitively designed to give the agent the choice of either \emph{exploring} the corpus by performing a broader search through the more permissive disjunctive query (documents are retrieved if either of the entities are present), or instead \emph{exploiting} particular regions of the corpus through the more restrictive conjunctive query (the documents must contain both entities). 

Because \emph{conjunctive} queries return documents with the text of both entities, they are more likely to identify relations connecting the entities.
%
However, there is also an increased risk that such queries will end up not finding any satisfying documents, especially when the entities are not closely related, resulting in waisting one iteration in the search process. 
On the other hand, 
\emph{disjunctive} queries are designed to return a larger set of documents, which, 
reduce the likelihood of returning an empty set. 
But they introduce the risk of processing more potentially irrelevant documents, and potentially introducing more irrelevant entities.
\emph{Singleton} queries represent a compromise between conjunction and disjunction. They are designed to expand the set of existing queries to the knowledge graph, which may in turn be along paths that connect the target entities, but retrieving documents related to just one entity, rather than two.

Every entity or pair of entities in the current knowledge graph is eligible to serve as a parameter in a query template. 
The challenge is to choose which entities paired with query template type are more likely to retrieve documents containing candidate paths, using only the domain-agnostic information present in the $KG$.


As the search process proceeds, the number of entities in the knowledge graph grows, in turn increasing the number of possible query actions that can be constructed. 
RL quickly becomes intractable as the state and action space grows.
We therefore perform a beam search to fix the cardinality of the action space to a constant size.
In particular, we use cosine similarity (for entity pairs) and average tf-idf scores (for single entities) to rank the entities that might participate in constructing query actions.
The agent then chooses among the top $n$ entities/pairs for each query template, thus bounding the total number of actions available to the agent to $3n$ different queries at each step.

We rank candidate entity pairs that might participate in query templates involving two entities by computing the cosine similarity of the vector representations of the entities.
We use the natural language expression representation of the named entities to construct a continuous vector representation. The vector representation of each entity is built by averaging the word embedding vectors\footnote{We used the pretrained GloVe model provided by spaCy at \url{https://spacy.io/models/en\#en_core_web_lg}} of the words in the text of the named entity description.
This similarity works as a proxy indicator of how related those entities are, under the intuition that entities that have similar embeddings are more likely to participate in relations.
	
For singleton entity queries, we use the average tf-idf score of the entity's natural language description for ranking. The tf-idf score of an entity is derived from averaging the tf-idf score of the individual terms in the entity's natural language description.  Each term's frequency value is based on the \emph{complete corpus}. Tf-idf scores are often used as a proxy measure of term importance (the term occurs selectively with greater frequency within some documents), so here the intuition is that entities with higher tf-idf scores may be associated with higher recall in the corpus.

Finally, there is an additional non-query action that is available in every step of the search: \emph{early stop}.
If the agent choses to stop early, the search process transitions to a final, unsuccessful state. This deprives the agent from successfully finding a path, 
but avoids incurring further cost of processing more documents in a possibly unfruitful search. 


\subsection{State Representation Features}\label{sec:state_representation}




At each step during search, the focused reading agent will select just one action to execute (a query action or early stop) based on the current search state. The agent makes this decision using a model that estimates for each action the expected long-term reward that can be achieved by taking that action in the current state. Here we describe the collection of features used to represent the current state, provided as input to the model. 



\begin{table}[hbt]
\small
  \centering
  \begin{tabular}{ll}
  	\toprule
    \multicolumn{1}{c}{\textbf{\textit{Category}}}& \multicolumn{1}{c}{\textbf{\textit{Feature}}}\\
    \midrule
    \multirow{4}{*}{\textit{Search state}}& Iteration number \\
    & Doc set size \\
    & \# of vertices in KG\\
    & \# of edges \\
    \midrule
    \multirow{2}{*}{\textit{Endpoints}} & Embedding of $E1$\\
    & Embedding of $E2$\\
    \midrule
    \multirow{2}{*}{\textit{Query}} & {\small Cosine sim. or avg tf-idf score}\\
    & \# of new documents to add\\
    \midrule
    \multirow{2}{*}{\textit{Topic Modeling}} & $\Delta$ Entropy of queries\\
    & KL Divergence of query\\
    \bottomrule
  \end{tabular}
  \caption{State representation features}
  \label{tab:statespace}
\end{table}

Table \ref{tab:statespace} provides a summary of the features included in the state representation. We group them into four categories.

\begin{itemize}
	\item \emph{Search state features}: Information about the current state of the search process including: the number of documents that have been processed so far; how long has the search been running (expressed in iterations); and the size of the knowledge graph. 
	\item \emph{Endpoints of the search}:  $E1$ and $E2$ represent the original target concepts that we are trying to find a path between. The identity of the endpoints determines the starting point of the search and conditions the theme of the content sought during the search. This information is provided to the model using the vector embedding representations of $E1$ and $E2$. 
	\item \emph{Query features}: 
	We include in the state representation features the score with which each of the $3n$ queries in the action space is ranked. This includes the cosine similarity for conjunction and disjunction queries and  tf-idf score for singleton queries. The intuition is that the score may be correlated with the expected long-term reward.
	For each query action, we will also see the identities of which documents will be retrieved. This allows us to count how many documents will be retrieved that have not already contributed to the $KG$, and this is included in the state representation for each action.	
	\item \emph{Topic modeling features}: 
	Finally, we would like to incorporate some indication of what information is contained in the documents that might be retrieved, and how it relates to the current entities within the $KG$. As a proxy for this information, we model the topics in the potentially retrieved documents, by peeking into the IR component to see the identity of documents that would be returned by the queries in the action space, and compare them to the topics represented in the $KG$ using two numerical scores: (a) an approximation of how broad or specific the topics are in the set of documents that would be returned by the query, and (b) an estimate of how the knowledge graph's \emph{topic distribution} would change if the query is selected as the next action. %
	
\end{itemize}



\subsection{Topic Modeling Features}\label{sec:topic_modeling}


Topic modeling features can be useful for staying \emph{on topic} throughout the search, avoiding drift into potentially irrelevant content.
Consider the following example: A user wants to know how the Red Sox and the Golden State Warriors are related by searching Wikipedia. 
While the two entities cover different sports in different regions of the United States, it is more likely that the connection will occur in a document about sports, e.g., they are both covered by the ESPN TV station.

We use Latent Dirichlet Allocation \cite{Blei2003LDA} (LDA) to provide the agent the ability to exploit topic information available in the corpus. LDA is unsupervised and requires only shallow processing of the corpus, namely, tokenizing and optionally stemming. This is essentially the same information required for constructing an inverted index for IR, so can be computed along with the IR component used in the focused reading system.


LDA produces a topic distribution for each document. 
We then aggregate the set of documents by summing the topic frequencies across documents and renormalizing. 
The topic distribution of the $KG$ is then the aggregation of the topic distributions of the documents processed so far in the search process. The topic distribution of a query is the aggregation of the topic distributions of the \emph{unseen} documents returned by the query. 

We consider two statistics for relating topic distributions: topic entropy and Kullback-Leibler (KL) divergence. 

Intuitively, the \emph{entropy} of a topic distribution is an estimate of how \emph{specialized} a document is, that is, how much it focuses on a particular set of topics. For example, a document that only talks about a specific sport will generally have a topic distribution where the mass is concentrated only on the particular topics of that sport, and therefore have a lower entropy than another document that discusses sports and business. 
%
Document sets with overall higher entropy are more likely to introduce information about more topics to the knowledge graph, and therefore produce more opportunities for new links between a broader set of entities.
Lower entropy queries focus on a narrower set of topics, and thus, may introduce links between a restricted set of entities. 
The difference in entropy expresses this intuition in relative terms. We introduce a feature, $\Delta$ Entropy, as the difference in entropy between documents retrieved by a candidate action and the documents the action retrieved in the previous step. Positive values indicate that the candidate query will generally expand the topics compared to those fetched the last step while negative values indicate more restricted topic focus.

$\Delta$ Entropy measures how concentrated the mass is, but it does not tell us {\em how} the distributions are different. Two document sets may have completely different topic distributions, yet have the same or similar entropy. Kullback-Leibler (KL) divergence~\cite{kullback1951information}, also known as relative entropy, helps measure how different two distributions are with respect to each other, even if they have the same absolute entropy.
To capture this information, we compute the KL divergence between the topic distribution in the new documents (retrieved by the new query) and the topic distribution of the knowledge graph. This estimates how different the information in the new query is relative to what has already been retrieved.




\subsection{Reward Function Structure}\label{sec:reward_structure}

The overall goal of the focused reading learning processes is to identify a policy that efficiently finds paths of relations between the target entities while minimizing the number of documents that must be processed by the IE component. 
To achieve this, we want the reward structure of the MDP to incorporate the tradeoff between the number of documents that have to be read (the reading cost) and whether the agent can successfully find a path between the entities. 
%
Equation \ref{eqn:reward} describes the reward function, where $s_{t}$ represents the current state and $a_t$ represents the action executed in that state.
\begin{equation}\small
	r(s_{t}, a_t) = \begin{cases} 
	S &\mbox{if } s_{t+1} \text{ is succesful sate} \\
	- c \times m & \text{if } m > 0 \\
	- e & \text{if } m = 0 \\
	\end{cases}\label{eqn:reward}
\end{equation}
A positive reward $S$ (for ``success'') is given when executing $a_t$ results in 
a transition to a state whose knowledge graph 
contains a path connecting the target entities.  
Otherwise, the search is not yet complete and the a cost is incurred for processing $m$ documents with machine reading on step $t$. The cost is adjusted by a hyper parameter $c$ that controls the relative expense of processing a single document. Note that there may be actions that return an empty document set, incurring no cost from reading, but still not making progress in the search. 
To discourage the agent from choosing such actions, the hyperparameter $e$ controls the cost of executing an unfruitful action that returns no new information. 
(Specific parameter values used in this work are presented in Table \ref{tab:hyperparams} of Section 5.)



%% file: content/experiments.tex
To evaluate the focused reading learning method, we introduce a novel dataset derived from the English version of Wikipedia.
Our dataset consist of a set of 369 multi-hop search problems, where
a search problem is consists of a pair of entities to be connected by a path of relations, potentially connecting to other entities along the path. 

The foundation of the dataset is a subset of 6,880 Wikipedia articles from the WikiHop \cite{welbl-etal-2018-constructing} corpus. We used Wikification~\cite{RRDA11,cheng-roth-2013-relational} to extract named entities from these documents and normalize them to the title of a corresponding Wikipedia article. 
%
Wikification does not perform relation extraction, so we lack gold-standard relations. To overcome this limitation, in this paper we induce a relation between entities that co-occur within a window of three sentences. Every relation extracted this way can be traced back to at least one document in the corpus.

We create a gold-standard knowledge graph using the induced entities and relations, and we sample pairs of entities connected by paths to create search problems for the dataset. Table \ref{tab:dataset} contains a break-down of the number of elements in each subset of the dataset. 

\begin{table}[hbt]
\small
\centering
  \begin{tabular}{lr}
  	\toprule
   \multicolumn{1}{c}{\textbf{\textit{Element}}} & \multicolumn{1}{c}{\textbf{\textit{Size}}}  \\
    \midrule
    Corpus &  6880 articles\\
    \midrule
	\multicolumn{2}{c}{\textit{Search Problems}}\\
    Training & 230 problems\\
    Development & 500 problems\\
    Testing & 670 problems\\
    \midrule
    Total & 1400 problems\\
    \bottomrule
  \end{tabular}
  \caption{Multi-hop search dataset details.}\label{tab:dataset}
\end{table}

We train an LDA model\footnote{We used the LDA implementation provided by gensim \url{https://radimrehurek.com/gensim_3.8.3/}.} and constructed an information retrieval inverted index over the collection of documents.\footnote{Code and dataset files are found at \url{https://ml4ai.github.io/OpenDomainFR/}}



We used the Advantage Actor Critic algorithm~\cite{mnih2016asynchronous} (A2C) to implement our reinforcement learning focused reading method.\footnote{Implemented using the rlpyt library hosted at \url{https://github.com/astooke/rlpyt}.}
%
%
%

A2C is an actor-critic method and we use a single neural network architecture to model the action policy (actor) as well as the state value function (critic). 
We use a single neural network architecture to implement the A2C actor-critic model.
The architecture consists of a fully-connected feed-forward neural network with four layers and two output \emph{heads}. 
The first output head represents the approximation of the \emph{action policy} (the actor) as a soft-max activation layer whose size is the cardinality of the action space. This approximates the probability distribution of the actions given the current state.
The second head approximates the \emph{state value}, as a single neuron with a linear activation. The state value estimates the expected long-term reward of using the estimated action distribution of the first head in the current state.
Altogether the model consists of approximately 3.79 million parameters.

Table \ref{tab:hyperparams} lists the hyper-parameter values used in our experiments.~\footnote{Hyper-parameter values were determined through manual tunning.}


\begin{table}[hbt]
\small
\centering
  \begin{tabular}{lr}
  \toprule
    \multicolumn{1}{c}{\textbf{\textit{Hyper-parameter}}}& \multicolumn{1}{c}{\textbf{\textit{Value}}} \\
    \midrule
    \multicolumn{2}{c}{\textit{Environment}}\\
    \# entities per query template & 15\\
    Maximum \# of steps & 10\\
    \midrule
    \multicolumn{2}{c}{\textit{Reward Function}}\\
    Successful outcome $S$ & 1000\\
    Document processing cost $c$ & 10\\
    Empty query cost $e$ & 100\\
    \midrule
    \multicolumn{2}{c}{\textit{Training}}\\
    Mini-batch size & 100\\
    \# Iterations & 2000\\
    \bottomrule
  \end{tabular}
  \caption{Hyper-parameter values}\label{tab:hyperparams}
\end{table}

\begin{table*}[hbt]
\centering
\begin{adjustbox}{max width=\textwidth}
  \begin{tabular}{ccccccc}
\toprule 
& &&& \multicolumn{3}{c}{\textbf{\textit{Average Steps}}}  \\
                   &   \textbf{\textit{Success Rate}} & \textbf{\textit{Processed Documents}} & \textbf{\textit{Documents per Success}} & \textbf{\textit{Overall}} & \textbf{\textit{Successes}} & \textbf{\textit{Failures}} \\
\midrule
\multicolumn{7}{c}{\textit{Baselines}}\vspace{.5em}\\
          \emph{Random} &   25.04 (0.014) &   56,187.8 (3,197.6) &        449.83 (34.34) &              8.41 (0.06) &                3.66 (0.25) &                    10 (0) \\
          \emph{Conditional} &   23.92 (0.008) &  49,609.8 (4,215.11) &        415.03 (36.01) &              8.51 (0.06) &                3.78 (0.07) &                    10 (0) \\
          \emph{Cascade} &   32.84 (0.01) &  62,058.2 (3,686.57) &       378.15 (23.87) &              7.42 (0.05) &                2.93 (0.19) &               9.61 (0.06) \\
          \midrule 
          	\midrule
\multicolumn{7}{c}{\textit{All Features}}\vspace{.5em}\\
\emph{Dropout 0.2} &     36 (0.007)* &    58,552.2 (719.67)* &        325.41 (8.43)* &              6.56 (0.05) &                2.22 (0.06) &               9.01 (0.04) \\
          \emph{Dropout 0.5} &  36.64 (0.004)* &       100,869 (4,121) &         550.76 (26.2) &                 7 (0.04) &                2.37 (0.08) &               9.67 (0.06) \\
           \emph{No Embs} &  26.3 (0.008) &       \textbf{39,433 (1,678.2)*} &         428.82 (19.51) &                 6.52 (0.03) &                2.37 (0.08) &               7.74 (0.07) \\
          \midrule 
          \midrule
\multicolumn{7}{c}{\textit{No Query Features}}\vspace{.5em}\\
\emph{Dropout 0.2} &   33.68 (0.003) &  42,022.2 (2,071.89)* &       \textbf{249.57 (13.02)*} &              4.56 (0.05) &                2.02 (0.03) &               5.84 (0.08) \\
          \emph{Dropout 0.5} &  36.48 (0.003)* &   62,126.6 (1,900.75) &       340.62 (10.79)* &              5.95 (0.06) &                2.28 (0.03) &               8.06 (0.09) \\
          \emph{No Embs} &   35.6 (0.005)* &  58,025.8 (1,085.72)* &        325.99 (4.26)* &              6.37 (0.07) &                 2.2 (0.07) &               8.68 (0.12) \\
          \midrule
\multicolumn{7}{c}{\textit{No Search Features}}\vspace{.5em}\\
\emph{Dropout 0.2} &  35.92 (0.002)* &    55,723 (1,437.01)* &        310.27 (8.13)* &              6.42 (0.04) &                2.15 (0.03) &               8.82 (0.07) \\
          \emph{Dropout 0.5} &  35.32 (0.003)* &  53,227.4 (1,429.88)* &        301.42 (8.77)* &              5.41 (0.07) &                2.09 (0.02) &               7.22 (0.11) \\
          \emph{No Embs} &  \textbf{37.16 (0.004)*} &   97,612.2 (4,550.54) &        525.48 (26.72) &                 6.92 (0) &                2.44 (0.08) &               9.56 (0.01) \\
          \midrule
\multicolumn{7}{c}{\textit{No Topic Features}}\vspace{.5em}\\
\textbf{Dropout 0.2} &  \textbf{35.56 (0.004)*} &  \textbf{51,757.4 (1,510.95)*} &        \textbf{291.11 (8.36)*} &              \emph{5.92 (0.02)} &                \textbf{2.15 (0.07)} &                  \textbf{8 (0.05)} \\
          \emph{Dropout} 0.5 &  35.72 (0.007)* &    55,637 (1,456.53)* &        311.58 (8.74)* &              5.56 (0.05) &                2.13 (0.06) &               7.46 (0.06) \\
           \emph{No Embs} &   28.52 (0.004) &  50,634.6 (2,060.88)* &         355.05 (12.1) &              6.74 (0.06) &                 3.5 (0.04) &                8.03 (0.1) \\
\bottomrule
\end{tabular}
\end{adjustbox}

  \caption{Feature sets ablation results. * denotes the difference w.r.t. the cascade baseline is statistically significant.}\label{tab:ablation}
\end{table*}

\begin{table*}[th]
  \centering
  \begin{adjustbox}{max width=\textwidth}
  \begin{tabular}{ccccccc}
  \toprule
  & &&& \multicolumn{3}{c}{\textbf{\textit{Average Steps}}}  \\
                     &   \textbf{\textit{Success Rate}} & \textbf{\textit{Processed Documents}} & \textbf{\textit{Documents per Success}} & \textbf{\textit{Overall}} & \textbf{\textit{Successes}} & \textbf{\textit{Failures}} \\
  \midrule
  \multicolumn{7}{c}{\textit{Baseline}}\vspace{.5em}\\
  \emph{Cascade} &  36.52 (0.008) &  83,252.4 (2,538.11) &       339.39 (13.01) &              7.18 (0.06) &                2.81 (0.05) &               9.69 (0.04) \\
  \midrule
  \multicolumn{7}{c}{\textit{No Topic Features}}\vspace{.5em}\\
  \emph{Dropout 0.2} &  39.02 (0.007)* &   \textbf{79,737.2 (1,664.65)} &       \textbf{304.22 (10.06)*} &              \textbf{6.28 (0.04)} &                2.16 (0.08) &               \textbf{8.91 (0.03)} \\
  \midrule
  \multicolumn{7}{c}{\textit{All Features}}\vspace{.5em}\\
  Dropout 0.2 &  \textbf{39.49 (0.003)*} &   85,637.8 (1,751.95) &         322.71 (7.99) &              6.34 (0.04) &                \textbf{2.16 (0.03)} &               9.07 (0.06) \\
  \bottomrule
  \end{tabular}
   \end{adjustbox}
    \caption{Results of the best model in the testing dataset. Quantities are averages over five runs with different random seeds and standard deviations are shown in parentheses.
    * denotes the difference w.r.t. the cascade baseline is statistically significant.}\label{tab:validation}
  \end{table*}

We performed an ablation analysis on the development dataset to find the best configuration of features. 

The development dataset contains five hundred search problems. The set of endpoints of the search problems does not overlap with those of the training and validation datasets. This is enforced to avoid any accidental leak of training information.

Table \ref{tab:ablation} contains the results of the ablation experiments. All the search problems were repeated five times with different random seeds. 
The key columns of the table are defined as follows. \emph{Success Rate} represents the percentage of problems in the test set for which the agent connected the endpoints. \emph{Processed Docs} provides the number of documents processed in all the search problems of the test set. \emph{Docs per Success} is a summary the other two columns: it contains the number of documents processed divided by the number of successes. This ratio is an aggregate statistic useful for comparing the performance between different policies.
We report the sample averages and their standard deviations in parentheses. For example, the \emph{Success Rate} column displays the average and standard deviation of five success rate calculations over five hundred search problems. The \emph{Processed Documents} column displays the average and standard deviation of the cumulative count of documents processed in the search problems, and so forth.


We implement three baseline policies that were not derived using RL:

\begin{itemize}
	\item \emph{Random}: Uniformly randomly selects a query from all possible queries constructed from eligible combinations of entities assigned to the query templates.
	\item \emph{Conditional Random}: Uniformly randomly selects a query template, \emph{conjunction, disjunction} and \emph{singleton} and then choses the uniformly randomly selects the entities to parameterize the template.
	\item \emph{Cascade}: Uniformly randomly samples a pair of entities and executes a conjunction query. If the result set does not contain any documents, then the agent selects a disjunction query with the same entities.
\end{itemize}

For consistency, each baseline was also evaluated with five different random seeds over the testing set. The top part of Table \ref{tab:ablation} shows the results of the baseline policies.

To test for statistical significance, we performed a non-parametric bootstrap resampling test with ten thousand samples for the the following metrics: \emph{success rate, processed documents} and \emph{documents per success}. For each metric, we calculated the difference between the result of the cascade baseline and the result of each of the reinforcement learning (RL) policies. If $p \leq 0.05$ of the difference being in favor of the reinforcement learning policy, the quantity is starred in the table.

In terms of success rate, most of the reinforcement learning models perform better than the cascade baseline. The notable exceptions are two feature configurations that do not use endpoint embeddings. These configurations are the one that considers all feature classes and the one that does not consider topic features.

Excluding query features from training produces models that process fewer documents per success with or without endpoint embeddings. 

Excluding search features produced in average models with higher success rate, with or without embeddings, but does so while processing more documents compared to other configurations.

The configurations that exclude query features produced models with the best numbers of documents per success. When the topic features are excluded, a similar result is achieved, but the number of documents per success of model that does not use endpoint embeddings is not statistically significantly lower than that of the cascade baseline.

Nonetheless the model that has the best balance between \emph{success rate} and \emph{documents per success} is the one that excludes topic features and trains with a dropout coefficient of $0.2$ on the endpoint embeddings. We  use this model to evaluate the validation dataset.

Table \ref{tab:validation} displays the results of the cascade baseline, which shows the strongest performance among the baseline policies, and the chosen reinforcement learning model on the validation dataset.
The validation dataset contains 650 search problems and the set of endpoints of its search problems is disjoint from the other datasets' for the same reason, to avoid leaking any training or development signal into the validation dataset. We did the same non-parametric bootstrap test for statistical significance.
The reinforcement learning policy achieves approximately $2.5\%$ higher average success rate on the testing dataset that the cascade baseline policy. While it also processes fewer documents in average, the difference is not statistically significant, but when considering the number of documents per success, the result is indeed significant, requiring approximately thirty five documents in average than cascade.

%% file: content/conclusions.tex
We proposed a focused reading methodology to automatically learn how to direct search in large corpora while iteratively building a knowledge base. The knowledge base is modeled as a graph, which in turn is used to focus the search toward documents that appear relevant. Our methodology complements existing information retrieval and machine tools. 
We evaluated focused reading on a set of search problems extracted from English Wikipedia and demonstrated that reinforcement learning with a state representation based on features about dynamics of the search process and the properties of the corpus is more effective and efficient than heuristic baselines.
In this methodology, inference in a knowledge graph acquired during the search process is agnostic of the semantics of the concepts and their relations. Their quality depends on the machine reading components used to extract them.

In future work, we plan to explore approaches for incorporating the semantics of the relations along the multi-hop paths that connect the target entities. 
Crucially, this includes incorporating additional constraints based on topic context.
Providing context to a search problem could prove useful to better focus the search process and to improve the accuracy of inference. We also plan to adapt the focused reading methodology to handle other class of search problems, e.g., slot filling tasks where the endpoints are underspecified.